\documentclass{article}

\usepackage[preprint, nonatbib]{neurips_2021}

\usepackage[utf8]{inputenc} 
\usepackage[T1]{fontenc}    
\usepackage{url}            
\usepackage{booktabs}       
\usepackage{amsfonts}       
\usepackage{nicefrac}       
\usepackage{microtype}      
\usepackage{threeparttable}
\usepackage{multirow}
\usepackage{graphicx}
\usepackage[square,numbers]{natbib}
\usepackage{amsmath}
\usepackage{mathrsfs}
\usepackage{color}
\usepackage{xcolor}
\usepackage{makecell}
\usepackage{listings}
\usepackage{wrapfig}
\usepackage{amsthm}

\usepackage{subcaption}
\usepackage{float}
\usepackage{epsfig}
\usepackage{bbm}
\usepackage{comment}
\usepackage{booktabs}
\usepackage{color}
\usepackage{multicol}
\usepackage[ruled,vlined,linesnumbered,noresetcount]{algorithm2e}
\usepackage{algpseudocode}

\newtheorem{theorem}{Theorem}
\newtheorem{definition}{Definition}

\newtheorem{innercustomthm}{Theorem}
\newenvironment{customthm}[1]
  {\renewcommand\theinnercustomthm{#1}\innercustomthm}
  {\endinnercustomthm}

\definecolor{deepred}{rgb}{0.631,0.102,0.102}
\usepackage[colorlinks,linkcolor=deepred]{hyperref}
\hypersetup{
     colorlinks = false,
     urlcolor = deepred
     }

\newcommand{\AlgName}{\textsc{DataSifter}\xspace}

\newcommand{\add}[1]{\textcolor{black}{#1}}

\newcommand{\dataset}{\mathcal{D}}
\newcommand{\uFunc}{U}
\newcommand{\R}{\mathbb{R}}
\newcommand{\M}{\mathcal{M}}
\newcommand{\I}{\mathcal{I}}
\newcommand{\floor}[1]{\left \lfloor{#1}\right \rfloor}
\newcommand{\ceil}[1]{\left \lceil{#1}\right \rceil}

\title{A Unified Framework for Task-Driven Data Quality Management}

\author{%
  Tianhao Wang \\
  Harvard University \\
  \texttt{tianhaowang@fas.harvard.edu} \\
  \And
  Yi Zeng \\
  Virginia Tech \\
  \texttt{yizeng@vt.edu} \\
  \And
  Ming Jin \\
  Virginia Tech \\
  \texttt{jinming@vt.edu} \\
  \And
  Ruoxi Jia \\
  Virginia Tech \\
  \texttt{ruoxijia@vt.edu} \\
}

\begin{document}
\maketitle

\begin{abstract}
High-quality data is critical to train performant \textit{Machine Learning}~(ML) models, highlighting the importance of \textit{Data Quality Management}~(DQM). Existing DQM schemes often cannot satisfactorily improve ML performance because, by design, they are oblivious to downstream ML tasks. Besides, they cannot handle various data quality issues (especially those caused by adversarial attacks) and have limited applications to only certain types of ML models. Recently, data valuation approaches (e.g., based on the Shapley value) have been leveraged to perform DQM; yet, empirical studies have observed that their performance varies considerably based on the underlying data and training process. In this paper, we propose a \emph{task-driven, multi-purpose, model-agnostic} DQM framework, \AlgName, which is optimized towards a given downstream ML task, capable of effectively removing data points with various defects, and applicable to diverse models. Specifically, we formulate DQM as an optimization problem and devise a scalable algorithm to solve it. Furthermore, we propose a theoretical framework for comparing the worst-case performance of different DQM strategies. Remarkably, our results show that the popular strategy based on the Shapley value may end up choosing the worst data subset in certain practical scenarios. Our evaluation shows that \AlgName achieves and most often significantly improves the state-of-the-art performance over a wide range of DQM tasks, including backdoor, poison, noisy/mislabel data detection, data summarization, and data debiasing. 

\end{abstract}

\section{Introduction}
\label{sec:intro}
High-quality data is a critical enabler for high-quality \emph{Machine Learning}~(ML) applications. However, due to inevitable errors, bias, and adversarial attacks occurring during the data generation and collection processes, real-world datasets often suffer various defects that can adversely impact the learned ML models. Hence, \textit{Data Quality Management}~(DQM) has become an essential prerequisite step for building ML applications. 

DQM has been extensively studied by the database community in the past. Early works~\cite{neutatz2021cleaning,chu2015katara,schelter2019unit} consider DQM as a standalone exercise without considering its connection with downstream ML applications. Studies have shown that such ML-oblivious DQM may not necessarily improve model performance~\cite{neutatz2021cleaning}; worse yet, it may even degrade model performance~\cite{amershi2019software}. More recent work started to tailor the DQM strategies to specific ML applications~\cite{krishnan2016activeclean,karlavs2020nearest,krishnan2019alphaclean}. Still, they apply only to simple models such as convex models, nearest neighbors, and specific data quality issues such as outlier detection. In parallel with these research efforts, the ML community has intensively investigated techniques focused on addressing a broad variety of data quality issues, such as adversarial~\cite{wang2019neural,chen2019deepinspect} and mislabeled~\cite{koh2017understanding,pruthi2020estimating} data detection, anomaly detection~\cite{du2019robust}, dataset debiasing~\cite{zemel2013learning,madras2018learning,wang2021learning}. However, a DQM scheme that can comprehensively remedy various types of data defects is still lacking.

Our paper aims to address the limitations of prior DQM schemes by developing a unified DQM framework with the following properties: (1) \emph{multi-purpose} -- to handle various data quality issues; (2) \emph{task-driven} -- to effectively utilize the information from downstream ML tasks; and (3) \emph{model-agnostic} -- to incorporate different ML models. The line of existing work closest to achieving these goals is what we will later refer to as data valuation-based approaches. 
These approaches first adopt some importance quantification metric, e.g., influence functions~\cite{koh2017understanding}, Shapley values~\cite{ghorbani2019data,jia2019towards} and least cores~\cite{yan2020ifyoulike}, to quantify each training point according to the contributions toward the training processes, then decide which data to retain or remove based on the valuation rankings. 
While some existing data valuation-based approaches satisfy the three desiderata, empirical studies have shown that their performance varies considerably based on the underlying data and the learning process. Moreover, there is no clear understanding of such performance variation or formal characterization of the worst-case performance.



In this paper, we start by formulating various DQM tasks into optimal data selection problems. The goal is to find a subset of data points that achieve the highest performance for a given ML task. We propose \AlgName, a multi-purpose, task-driven, model-agnostic DQM framework that first learns a data utility model from a small validation set, then selects the subset of data points by optimizing the acquired utility model. With the acquired data utility model, \AlgName can go beyond the functionalities offered by existing DQM schemes and further estimate the utility of selected data points. Such information could help data analysts to decide how many data points to choose or whether there is a need to acquire new data. Furthermore, we present a novel theoretical framework based on domination analysis which allows one to rigorously analyze the worst-case performance of data valuation-based DQM approaches and compare them with our approach. Specifically, we show that data valuation-based DQM approaches have unsatisfying worst-case performance guarantees. In particular, the popular Shapley value-based approach will select the worst data in some commonly occurring scenarios. We conduct a thorough empirical study on a range of ML tasks, including adversarially perturbed data detection, noisy label/feature detection, data summarization, and data debiasing. Our experiments demonstrate that \AlgName achieves and most often significantly improves the state-of-the-art performance of data valuation-based approaches on various tasks.

\section{Related Work}
The major differences between this paper and the related works are summarized in Table~\ref{table:summary-of-properties}.

\begin{wraptable}{R}{0.59\linewidth}
\centering
\scriptsize
\begin{tabular}{@{}l|p{0.85cm}<{\centering}p{0.85cm}<{\centering}p{0.85cm}<{\centering}p{0.85cm}<{\centering}p{0.85cm}<{\centering}@{}}

\multirow{2}{*}{Method Type} & \textbf{Multi-} & \textbf{Task-}  & \textbf{Model-} & \textbf{Est.} \\
& \textbf{purpose} & \textbf{Driven}  & \textbf{Agnostic} & \textbf{Utility}\\
\hline
\textbf{Traditional}         & $\times$               & $\times$                      & $\times$                & $\times$                            \\
\textbf{Data Cleaning}           & $\times$               & $\circ$                      & $\circ$                & $\times$                            \\
\textbf{Perm-Shapley \cite{maleki2015addressing}}            & $\checkmark$           & $\checkmark$               & $\checkmark$            & $\times$                            \\
\textbf{TMC-Shapley \cite{ghorbani2019data}}             & $\checkmark$           & $\checkmark$                & $\checkmark$            & $\times$                            \\
\textbf{G-Shapley \cite{ghorbani2019data}}               & $\checkmark$           & $\checkmark$                 & $\times$                & $\times$                            \\
\textbf{KNN-Shapley \cite{jia2019efficient}}             & $\times$               & $\times$                 & $\checkmark$            & $\times$                            \\
\textbf{Least Core \cite{yan2020ifyoulike}}              & $\checkmark$           & $\checkmark$             & $\checkmark$            & $\times$                            \\
\textbf{Leave-one-out \cite{koh2017understanding}}           & $\checkmark$           & $\checkmark$                & $\checkmark$            & $\times$                            \\
\textbf{Infl. Func. \cite{koh2017understanding}}      & $\times$               & $\checkmark$              & $\times$                & $\times$                            \\
\textbf{TracIn \cite{pruthi2020estimating}}                  & $\times$               & $\checkmark$             & $\times$                & $\times$                            \\
\textbf{\AlgName} & $\checkmark$           & $\checkmark$              & $\checkmark$            & $\checkmark$                        \\ 
\Xhline{1pt}
\end{tabular}
\caption{Summary of the differences between previous works with our methods (\AlgName). $\circ$ means only some of the techniques in the type satisfy the property.
}
\label{table:summary-of-properties}
\end{wraptable}

\textbf{Data Cleaning.} Classical data cleaning methods are based on simple attributes of a dataset such as completeness~\cite{schelter2019unit}, consistency~\cite{kandel2011wrangler}, and timeliness~\cite{chu2015katara}; however, these attributes may not necessarily correlate with the actual utility of data for training machine learning models. Recent works leverage the information about downstream ML tasks to guide the cleaning process.
ActiveClean \cite{krishnan2016activeclean} explored task-driven data cleaning for convex models by selecting data points for human screening. BoostClean \cite{krishnan2017boostclean} sought to automate the manual cleaning by determining a predefined cleaning strategy from a library using boosting. AlphaClean \cite{krishnan2019alphaclean} also aimed to automate the cleaning process but relied on parallelized tree search. However, those framework's efficacy and generalizability are limited by the cleaning library. Furthermore, the recursive nature of the automatic selection process constrained the use case of those methods to only small models and datasets. CPClean \cite{karlavs2020nearest} proposed a different strategy for nearest neighbor models based on the concept of Certain Prediction. Still, CPClean is designed explicitly for SQL datasets with greedy repairing, making it difficult to generalize to larger-scaled cases like image datasets. In summary, the state-of-the-art data cleaning methods are only applicable to certain classes of ML models and datasets. Besides, adapting those cleaning works to other domains requires manual recollection of the cleaning library or human intervention, which can be impractical in many cases.

\textbf{Data Importance Quantification.} 
One simple idea to quantify data importance is to use the leave-one-out error. \cite{koh2017understanding} provides an efficient algorithm to approximate leave-one-out error for each training point. Recent works leverage credit allocation schemes originated from cooperative game theory to quantify data importance. Particularly, Shapley value has been widely used \citep{ghorbani2019data, jia2019towards, jia2019efficient, jia2019scalability, wang2020principled}, as it uniquely satisfies a set of desirable axiomatic properties. More recently, \cite{yan2020ifyoulike} suggests that the Least core is also a viable alternative to Shapley value for measuring data importance. However, computing the exact Shapley and Least core values are generally NP-hard. Several approximation heuristics, such as TMC-Shapley \citep{ghorbani2019data}, G-Shapley \citep{ghorbani2019data}, KNN-Shapley \citep{jia2019efficient}, have been proposed for the Shapley value. Despite their computational advantage, they are biased in nature. On the other hand, unbiased estimators such as Permutation Sampling \citep{maleki2015addressing} and Group Testing \citep{jia2019towards} still require retraining models many times for any descent approximation accuracy. TracIn \cite{pruthi2020estimating} estimates the importance by tracing the test loss change caused by a training example during the training process. The representer point method~\cite{yeh2018representer} captures the importance of that training point by decomposing the pre-activation prediction of a neural network into a linear combination of activations of training points. Many of the aforementioned works can only be applied to differentiable models.

\section{Formalism and Algorithmic Framework}
\label{sec:method}



In general, DQM aims to find a subset of data points with the highest utility. We use the data utility function to characterize the mapping from a set of data points to its utility.
Formally, given a dataset $\dataset$ of size $n$, a \emph{data utility function} $\uFunc: 2^\dataset \rightarrow \R$ maps a set of data points $S \subseteq \dataset$ to a real number indicating the performance of the ML model trained on the set, such as test accuracy and fairness.



With the notion of the data utility function, one can abstract DQM tasks as a \emph{data selection problem}: 
$\max_{|S|=k}\uFunc(S)$, where $k$ indicates the selection budget with $0<k<n$, which can be predetermined (e.g., based on the prior knowledge about potential data defects or computational requirements). Moreover, the DQM tasks without a specific selection budget can be reduced to a sequence of data selection problems with different values of $k$.

With the abstraction above, one straightforward way to optimally select data is to exhaustively evaluate $\uFunc(S)$ for all possible size-$k$ subsets $S \subseteq \dataset$ and choose the one that achieves the highest utility. 
Of course, this naive algorithm requires prohibitively large computational resources because the number of utility evaluations is exponential in $k$, and worse yet, each evaluation of data utility function requires retraining the model. 
Fortunately, recent work shows that many common data utility functions can be effectively learned with a relatively small amount of samples \citep{wang2021one} because they are ``approximately'' submodular~\cite{balcan2011learning}. 
The ``approximate submodularity'' property allows efficient maximization of data utility functions through simple greedy algorithms~\citep{minoux1978accelerated, horel2016maximization, hassidim2018optimization, das2018approximate, chierichetti2020on}. 
Hence, combining data utility function learning and greedy search enables an efficient algorithm for data selection problems. 

Specifically, we extend and generalize the data utility learning and optimization technique originally proposed in \citep{wang2021one} for active learning to DQM. The proposed DQM framework, termed \AlgName, proceeds in two phases: \emph{learning} and \emph{selection} phase. 




\begin{figure}[t]
	\centering
	\includegraphics[width=\linewidth]{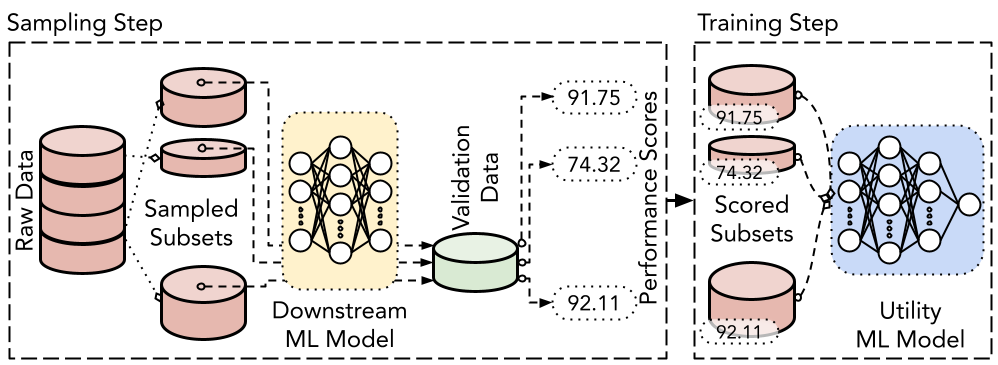}
	\caption{Overview of the learning phase, which consists of two functional steps, the utility sampling step and the training step of the utility ML model. We randomly select subsets from the raw data during the sampling step and assign a utility score on each set based on the downstream ML model's performance evaluated over a validation set. Then, we train the utility ML model that uses those scored pairs to  acquire the ability to predict a performance score for a collection of data.}
	\label{fig:flow-chart}
\end{figure}

\textbf{Learning Phase. }
Figure \ref{fig:flow-chart} depicts the learning phase of the \AlgName, which consists of a utility sampling step and a utility model training step. In particular, we assume that we have access to a small validation set representative for potential test samples. Thus, the utility of any given subset can be estimated by feeding the model with this subset then evaluating its performance over the validation set. 
In the utility sampling step, we  randomly sample subsets of the training set, estimate the utility of each sampled subset, and label each using its utility score. We will refer to the scored subset as \emph{utility samples} hereinafter. To accelerate this step for large models such as deep nets, a small proxy model (such as logistic regression) can be used for approximating the utility since data utilities evaluated on deep nets and logistic regression are positively correlated, as shown in~\cite{wang2021learning}.
In the utility model training step, we learn a parametric model for the data utility function using the utility samples; particularly, our experiments adopt DeepSets~\citep{zaheer2017deep} as the utility model. For a large dataset, the utility sampling step could be conducted on a small portion of the dataset. Our empirical studies show that the learned utility model can still extrapolate the utility for the unseen part of the dataset. 




\begin{wrapfigure}{R}{0.46\linewidth}
    \centering
	\includegraphics[width=0.65\linewidth]{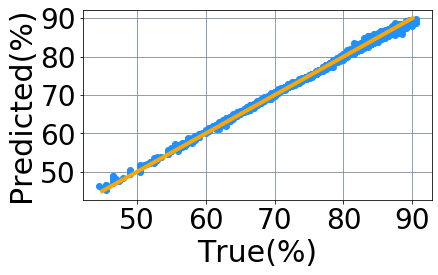}
	\caption{Predicted vs. True Utility for unseen subsets of logistic regression classifier trained on a synthetic dataset. Details are presented in supplementary materials.}
	\label{fig:utility-accuracy}
\end{wrapfigure}

\textbf{Selection Phase. }
We select high-quality data through optimizing the learned model for data utility functions obtained from the previous phase; specifically, we adopt a linear-time stochastic greedy algorithm~\cite{mirzasoleiman2015lazier} to perform optimization. 



Clearly, \AlgName is an optimal solution to the data selection problem if the validation data matches the test data exactly and there are no computational constraints. In practice, despite limited validation dataset and limited computational resources, \AlgName is still very effective in selecting high-quality data or filtering bad data as we will show in the evaluation section. 
In addition, with the learned data utility model, \AlgName can provide an estimate of the utility for the selected dataset (see example in Figure~\ref{fig:utility-accuracy}), which will be useful for data analysts to decide the number of data points to select.

\section{Worst-Case Analysis}
\label{sec:theory}

This section presents a theoretical framework for comparing the worst-case performance between \AlgName and data valuation-based DQM schemes, such as \emph{leave-one-out}~(LOO), Shapley value, and Least core\footnote{Least core may not be unique. In this paper, when we talk about the least core, we always refer to the least core vector that has the smallest $\ell_2$ norm, following the tie-breaking rule in the original literature \citep{yan2020ifyoulike}.}, and we assume no computational constraints.


We start by abstracting a general notion from data valuation-based DQM schemes in the literature. We call an algorithm that returns $S \subseteq \dataset$ of size $k$ a \emph{heuristic} to a (size-$k$) data selection problem on $\dataset$. The typical pattern of data valuation-based heuristics is that they first rank the data points according to their corresponding data importance metric and then prioritize the points with the highest importance scores. We will define the heuristics matching this selection pattern as \emph{linear heuristics}.

\begin{definition}[Linear heuristic]
\label{def:linearheuristic}
We say $\M$ is a linear heuristic for data selection problem if for every instance $\I = (\dataset, \uFunc)$, $\M$ works as follows: 
\begin{enumerate}
    \item Assign a score $v = (v_1, \ldots, v_n)$ for every data point $i \in \dataset$. 
    \item Sort $\dataset$ in the descending order according to $v$ and obtain sorted data sequence $\dataset'$. Certain rules are applied to break tie. 
    \item For any query of selecting $k$ high-quality data points, return the first $k$ data points in $\dataset'$. 
\end{enumerate}
\end{definition}


Our theoretical framework for studying the worst-case performance of data selection heuristics extends the domination analysis initially proposed in \cite{glover1997travelling}. Our worst-case performance metric is \emph{domination number}, which measures how many subsets achieve lower utility than the selected set in the worst-case scenario. 

\begin{definition}[Domination number]
The domination number of a heuristic $\M$ for the data selection problem is the maximum integer $d(n, k)$ s.t., for every problem instance $\I = (\dataset, \uFunc)$ on a dataset $\dataset$ of size $n$ and utility function $\uFunc$, $\M(\I, k)$ produces a size-$k$ subset $S \subseteq \dataset$ which has utility $\uFunc(S)$ no worse than at least $d(n, k)$ size-$k$ subsets. 
\end{definition}

The domination number is well defined for every data selection heuristic. A heuristic with a higher domination number may be a better choice than a heuristic with a smaller domination number due to the better worst-case guarantee. The best heuristic for data selection has domination number $d(n, k) = {n \choose k}$ for every $k \le n$, which means that it will select the size-$k$ data subset with the highest utility for every possible data utility function. 

Clearly, assuming no computational constraints, \AlgName is among the best heuristics which achieve the largest possible domination number. 
In contrast, the following result shows that no linear heuristic is the best whenever $n \ge 3$. We will defer all proofs to Appendix. 
\begin{theorem}
\label{thm:nooptimal}
For $n \ge 3$, there exists no linear heuristic $\M$ s.t.  $d(n, k) = {n \choose k}$ for every $k \in \{1, \ldots, n\}$. 
\end{theorem}

Furthermore, we can tighten the upper bound of the domination number for data valuation-based heuristics by noticing another common property: 
two data points will receive the same importance score if they contribute equally to all possible subsets of the training data. This property is often referred to as \emph{symmetry axiom} in the literature.


\begin{definition}[Symmetry axiom]
\label{def:symmetry}
We say a linear heuristic $\M$ satisfies symmetry axiom if its scoring mechanism satisfies: 
$
[(\forall S \in \dataset \setminus \{i, j\}) U(S \cup\{i\})=U(S \cup\{j\})] \Longrightarrow v_{i}=v_{j}
$.
\end{definition}

\begin{wrapfigure}{R}{0.33\linewidth}
    \centering
    \includegraphics[width=0.29\textwidth]{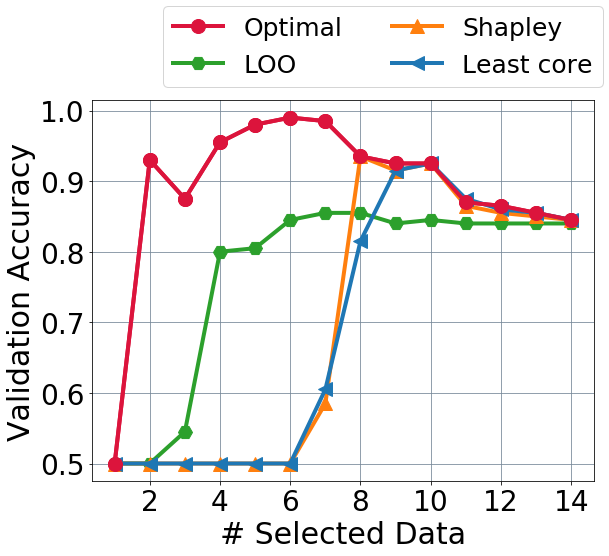}
    \caption{Results of data selection with different heuristics on a tiny dataset with natural redundancy. Dataset and implementation are detailed in the Appendix.}
    \label{fig:theory}
\end{wrapfigure}

The symmetry axiom may be desired for application scenarios requiring fairness, e.g., data importance scores are used to assign monetary rewards for data sharing or responsibility for ML decisions. However, for data selection, symmetry axiom may be undesirable because simply gathering high-value data points may lead to a set of redundant points. 
Based on this intuition, the following theorem gives an upper bound of domination number for non-trivial linear heuristics that with symmetry property. 
\begin{theorem}
\label{thm:symmetrybound}
If a linear heuristic $\M$ assigns different scores to different data points
and satisfies symmetry axiom, then the domination number of $\M$ is $d(n, k) \le c{\ceil{n/c} \choose k}$ where $c = \left \lfloor{\frac{n}{k}}\right \rfloor$.
\end{theorem}



To better illustrate the issue raised by symmetry axiom, we evaluate the LOO, Shapley, and least core heuristic on a synthetic dataset with 15 training data points (so that we can compute the exact Shapley and least core values, as well as obtain the optimal solution for data selection problem). The utility metric is the test accuracy of a \emph{Support Vector Machine}~(SVM) classifier trained on the dataset. We simulate the natural redundancy in a dataset by replicating 5 data points three times and adding slight Gaussian noise to differentiate. Figure \ref{fig:theory} shows that with small selection budgets, the subsets selected by all the heuristics have low utility as the heuristics fail to promote diversity during selection.

Notably, we show that the Shapley value heuristic would select the data subset with the lowest utility for certain data utility functions, including submodular ones. 
The Shapley value of a training point is calculated by taking a weighted average of the contribution of the point to all possible subsets of the training set, and the weights are independent of the selection budget $k$. 
Moreover, the Shapley value of training data weights higher for its marginal contributions on small datasets. Thus, data points that make a larger contribution on tiny datasets may be assigned with higher Shapley value, even if they make little or negative contributions in every dataset of desired selection size $k$. 


\begin{theorem}
\label{thm:shapleybound}
For any $n \ge 4$ and $k \in \{1, \ldots, n\}$, the domination number of Shapley value is $d(n, k) = 1$, even if the utility function $U$ is submodular. 
\end{theorem}

\begin{wraptable}{R}{0.5\linewidth}
\scriptsize
\centering
\begin{tabular}{@{}p{0.05cm}l|l|l@{}}
\Xhline{1pt}
\textbf{} & \multirow{2}{*}{\textbf{Task}} & \multicolumn{2}{c}{\textbf{Datasets}}\\
\cline{3-4}
 & & Main Text & Appendix\\
\Xhline{1pt}
\textbf{I.} & Backdoor Detection             & CIFAR-10  \citep{krizhevsky2009learning}&  MNIST\citep{lecun1998mnist}                       \\
\hline
\textbf{II.} & Poisoned Data Detection             & CIFAR-10 \citep{krizhevsky2009learning}&  Dog vs. Cat \citep{dogcat} \\
\hline
\textbf{III.} & Noisy Feature Detection             &CIFAR-10 \citep{krizhevsky2009learning}  &  MNIST \citep{lecun1998mnist}                  \\
\hline
\textbf{IV.} & Mislabeling Detection               & SPAM \citep{shams2013classifying}& CIFAR-10 \citep{krizhevsky2009learning}                 \\
\Xhline{1pt}
\textbf{V.}& Data Summarization                  & PubFig83 \citep{pinto2011scaling}& COVID-CT \citep{zhao2020COVID-CT-Dataset}               \\
\hline
\textbf{VI.} & Data Debiasing                  & Adult \citep{uci-dataset}& COMPAS \citep{yoon2018machine}                         \\
\Xhline{1pt}
\end{tabular}
\caption{Summary of DQM tasks and datasets. 
We discuss one dataset for each task and defer the results over the other dataset to the Appendix. 
}
\label{tb:summary-of-experiments}
\end{wraptable}

\section{Evaluation}
\label{sec:eval}
We evaluate \AlgName on six DQM tasks, as listed in Table \ref{tb:summary-of-experiments}.
We consider various benchmark models and datasets used in past literature for each DQM task. Since we can observe similar results on different datasets, this section will only describe the result on \emph{one} representative dataset for each task and leave the other dataset in the Appendix. Finally, we discuss the scalability of the \AlgName on larger datasets. The implementation details and the additional results are presented in the Appendix.  

\subsection{Baselines}

We focus on comparing data valuation-based approaches as they are closest to achieving the properties of multi-purpose, task-driven, and model-agnostic. 
We omit the data cleaning methods from the comparison as their applicability is limited to specific DQM tasks and specific models.
Specifically, we consider the following eight state-of-art data valuation-based approaches:
(1) \textit{Shapley Permutation Sampling}~\textbf{(Perm-SV)} \citep{maleki2015addressing}, a Monte Carlo-based algorithm for Shapley value estimation. 
(2) \textit{TMC-Shapley}~\textbf{(TMC-SV)} \citep{ghorbani2019data}, a refined version of the Perm-SV, where the computation is focused on the subsets whose utility changes significantly when an extra point is added.
(3) \textit{G-Shapley}~\textbf{(G-SV)} \citep{ghorbani2019data}, which approximates the Shapley value by anticipating the utility change caused by an extra point with its gradient.
(4) \textit{KNN-Shapley}~\textbf{(KNN-SV)} \citep{jia2019efficient}, which approximates the Shapley value by using the K-Nearest-Neighbor as a proxy model. 
(5) \textit{Least Core}~\textbf{(LC)} \citep{yan2020ifyoulike}, another data value notion in cooperative game theory.
(6) \textit{Leave-one-out}~\textbf{(LOO)} \citep{ghorbani2019data} evaluates the change of model performance when a data point is removed.
(7) \textit{Influence Function}~\textbf{(INF)} \citep{koh2017understanding}, which approximates the LOO error with influence functions.
(8) \textbf{TracIn} \citep{pruthi2020estimating}, which traces the test loss change during the training process whenever the training point of interest is utilized. 
(9) \textbf{Random} is a setting where we randomly select a subset from the target dataset. 

\add{For fair comparison between \AlgName and baselines, we fix the number of utility sampling as 4000 for \AlgName and baseline algorithms that require utility sampling. }
The implementations of \AlgName and baseline algorithms will be detailed in the Appendix. 
We repeat model training ten times for each selected set of data points to obtain the error bars.

\subsection{Results}

\subsubsection{Filtering out Harmful Data}

\begin{figure}
    \centering%
	\includegraphics[width=0.99\linewidth]{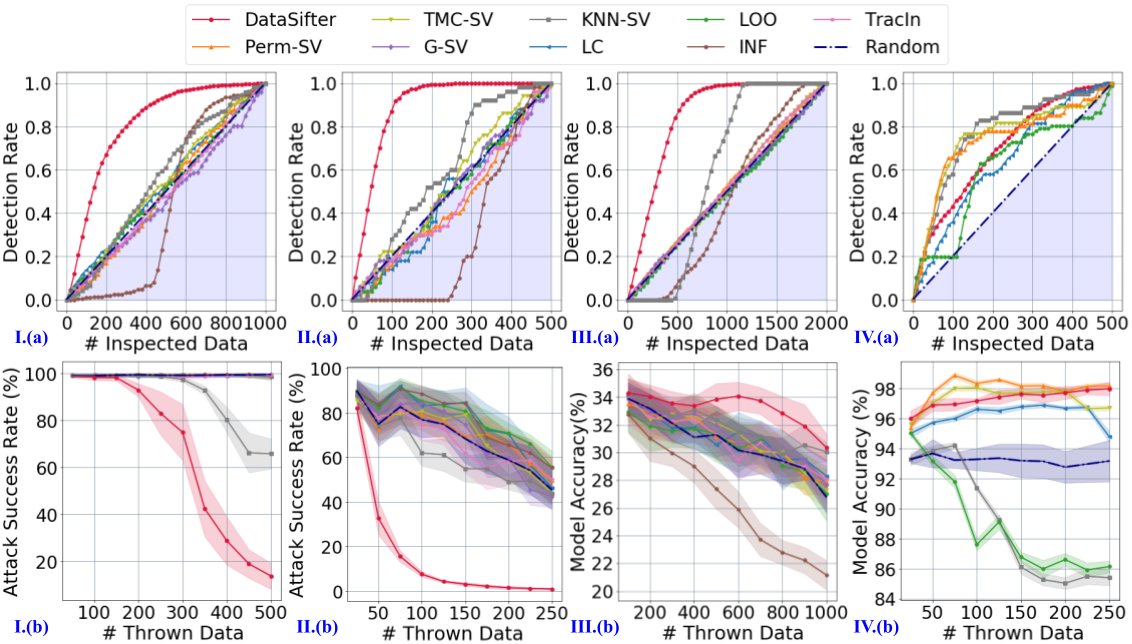}
	\caption{The experimental results and comparisons of the \AlgName under the case of filtering out harmful data (application I-IV). 
	The light blue region in each (a) graph represents the area that a method is no better than a random selection. For I.(b) and II.(b), we depict the Attack Success Rate (ASR), where a lower ASR indicates a more effective detection. For III.(b) and IV.(b), we show the model test accuracy, where a higher accuracy means a better selection.}
    \label{fig:all_detection}
\end{figure}

\label{sec:remoe-harmful}

Training data could be contaminated by various harmful examples, e.g., backdoor triggers, poison information, noisy/mislabeled samples. 
Our goal here is to identify data points that are most likely to be harmful. These points can either be discarded or presented with high priorities to human experts for manual cleaning.
To evaluate the performance of different DQM techniques, we examine the training instances according to the quality ranks outputted by each method and plot the change of the fraction of detected corrupted data with the fraction of the checked training data. 
Additionally, for poisoned/backdoor data detection, we plot the change of \textit{Attack Success Rate}~(ASR), and for noisy feature/label detection, we plot the change of model accuracies after filtering out the low-quality data points selected by each technique. 
\add{The validation data in utility sampling are 300 clean data points sampled from the test data of the corresponding datasets. }

\textbf{I. }\textbf{Backdoor Detection.}
Backdoor attacks~\citep{zeng2021rethinking,gu2017badnets} embed an exploit at training time that is subsequently invoked by the presence of a ``trigger'' at test time. They are considered particularly dangerous since they make models predict a target output on inputs with predefined triggers while still retain state-of-the-art performance on the clean data. Since data points with the backdoor triggers contribute little to the learning of clean validation samples, we could expect to identify them by minimizing the data utility model. This experiment studies the effectiveness of \AlgName for removing backdoored examples. 
We evaluate BadNets \cite{gu2017badnets} and Trojan attack \cite{liu2017trojaning}, the two most famous backdoor attacks in the literature. We adopted a three-layer CNN as the target model, a poison rate of 0.2, and a target label `Airplane.' 
Figure \ref{fig:all_detection} I.(a) and I.(b) elaborate the Trojan attack detection results for 
a 1,000-size randomly selected subset of the CIFAR-10 dataset. 
As we can see, \AlgName significantly outperforms other DQM approaches; for instance, it achieves a detection rate of 90\% with \textbf{51.17\%} fewer inspected data points than the others.

\textbf{II. Poisoned Data Detection.}
Adversaries make slight modifications to some training samples in data poisoning attacks to cause malicious behaviors in the test phase (e.g., misclassifying target test examples).
We evaluate different DQM techniques on two popular attacks, namely, feature collision attack \cite{shafahi2018poison} and influence function-based attack \cite{koh2017understanding}. These two are clean-label poisoning attacks where the attacker does not need to control the labeling of training data. We left the detailed descriptions of the attacks in the Appendix. 
Figure \ref{fig:all_detection} II.(a) and II.(b) show the results for feature collision attack \cite{shafahi2018poison} on a 500-size randomly selected CIFAR-10 subset, where 50 data points of class `cat' are perturbed with features extracted from a `frog' sample in the test set. We see that \AlgName significantly outperforms all other DQM methods in the poisoned data detection task; for instance, it attains a 90\% detection rate with \textbf{75.41\%} fewer examined data points. 



\textbf{III. Noisy Feature Detection.} 
Noise in features originated from sampling or transmitting (e.g., Gaussian noise) may decrease classification accuracy. Following the settings in \cite{wang2021one}, we add white noise to clean samples, and we evaluate the performance of each DQM technique on detecting those samples. For the CIFAR-10 dataset, we corrupt 25\% of the train data images by adding white noise. Based on Figure \ref{fig:all_detection} III.(a) and III.(b), we can conclude that \AlgName significantly outperforms all other methods on this task; for example, it achieves a 90\% of detection rate by examining \textbf{67.25\%} fewer data points. 
Meanwhile, the KNN-SV approach exhibits a distinctive trend -- it only starts finding the noisy data points until filtering out a certain amount of clean data. This is mainly because all noisy data points are out-of-distribution (OOD). The mechanism of KNN-SV tends to assign 0 values to OOD data points, while it will also assign negative values to some clean data points. We provide a more detailed explanation in the Appendix. 



\textbf{IV. Mislabeling Detection.}
Labels in the real world are often noisy due to automatic or non-expert labeling. Following \cite{ghorbani2019data, jia2019scalability}, we perform experiments on two datasets and present the results of SVM trained on Enron1 SPAM dataset \citep{shams2013classifying} and the CIFAR-10 dataset. We adopt a bag-of-words representation of the Enron1 for training. The noise flipping ratio is 15\%. Under this setting, Influence-based techniques and G-SV are not applicable since they require the model trained with gradient-based approaches. 
Figure \ref{fig:all_detection} IV.(a) and IV.(b) show that although \AlgName does not attain the highest detection rate, the accuracies of the model trained on the selected data are competitive with the most effective approaches.
For the Enron SPAM dataset, a small amount of mislabeled data points do not significantly affect the model performance; thus, those mislabeled samples could evade our detection based on the validation performance. By comparing Figure \ref{fig:all_detection} IV.(a) and IV.(b), we can tell such evasion is acceptable as the model trained over the data points selected by \AlgName still achieves a competitive accuracy. On the other hand, we find KNN-SV and LOO can accomplish a decent detection rate but end up with a lower validation accuracy. This is because they select very unbalanced data points, as both of them satisfy the symmetry axiom discussed in Section \ref{sec:theory}.

\begin{wrapfigure}{R}{0.65\textwidth}
    \centering
    \includegraphics[width=0.65\textwidth]{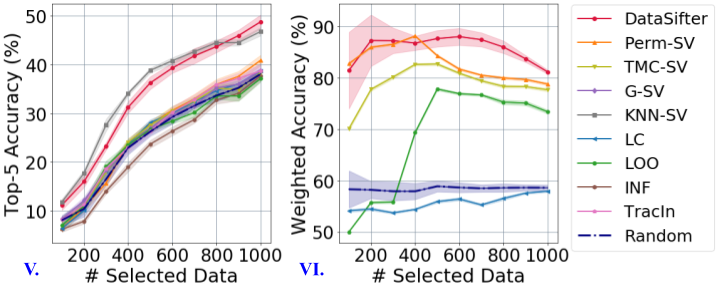}
    \caption{The experimental results and comparison of the \AlgName under the case of selecting high-quality data (application V and VI). 
    We depict the validation accuracy for both cases. A higher accuracy indicates a better performance. 
    }
    \label{fig:all-selection}
\end{wrapfigure}

\subsubsection{Selecting High-quality Data}

The DQM tasks considered in this section aim to select a subset that is most likely to help improve model test accuracy and fairness. 



\textbf{V. Data Summarization.} 
Data summarization aims to select a small, representative subset from a massive dataset, which can retain a comparable utility to that of the whole dataset. 
We use a convolutional neural network trained on the PubFig83 dataset in this experiment. Figure \ref{fig:all-selection} V shows that \AlgName and KNN-SV significantly outperform all the other DQM techniques, which have similar performance as the random selection. 


\textbf{VI. Data Debiasing. }
We explore whether DQM techniques can help select a subset of training data that improves both fairness and performance for the ML task. We use logistic regression trained on the UCI Adult Census dataset as the task model. We measure the fairness by weighted accuracy -- the average of model classification accuracy over females and that over males. 
G-SV, KNN-SV, and Influence-based techniques are not applicable for this application since they either require the model trained using the SGD, or are designed for computing data importance when the metric is test accuracy or loss. 
Therefore, we only compare with the remaining six baselines. Figure \ref{fig:all-selection} VI shows that \AlgName achieves the top-tire performance along with the Perm-SV. 

\begin{wrapfigure}{R}{0.65\textwidth}
    \centering
    \includegraphics[width=0.65\textwidth]{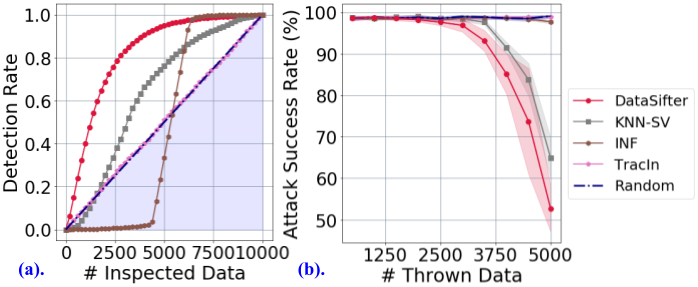}
    \caption{
    The experimental results and comparison of the \AlgName and baseline algorithms for detecting backdoored data on larger datasets. 
    }
    \label{fig:large}
\end{wrapfigure}

\subsection{Comparisons on Larger Datasets}
\label{sec:largedata}

We compare the scalability between \AlgName and other baselines on large datasets. We show the results for backdoor detection on a 10,000-size Trojan square poisoned CIFAR-10 subset here. 
For \AlgName, we only sample data subset utilities from 1000 data points as we did in Section \ref{sec:remoe-harmful} I, but use the learned utility model to select data points on the entire 10,000 data points. When executed on NVIDIA Tesla K80 GPU, the clock time for the utility sampling step is within 5 hours for 4000 utility samples with a small CNN model, as the data size is fairly small. The LOO, the Least core, and all the Shapley value-based approaches except KNN-SV did not terminate in 24 hours, so we remove them from comparison. As we can see from Figure \ref{fig:large}, \AlgName once again outperforms all the remaining approaches. The results show that the learned utility model can also provide utility estimations for a set of unseen data points, which largely improves the scalability of \AlgName. On the contrary, the existing valuation-based approaches cannot predict the importance of unseen data points. Thus their utility sampling has to be conducted over the entire dataset. 

\section{Limitations of the \AlgName}
\label{sec:limitations}
We now discuss two limitations of the \AlgName. 

\paragraph{Scalability.}
While Section \ref{sec:largedata} shows that \AlgName is often much more efficient than most of the other DQM schemes with similar design goals, its scalability to large dataset still needs further investigation. \AlgName could be slow as the utility sampling step requires retraining models for thousands of times. Although we already have several solutions for improving the scalability, such as using a smaller proxy model and/or only conduct utility sampling on a small portion of the dataset, it might still require hours of training. Further improving the scalability of \AlgName through some efficient approximation heuristics of data utility functions would be interesting future works. 

\paragraph{Utility Learning Model. }
In this work, we use the popular set function learning model--DeepSet--as our utility learning model for all of the experiments. However, as shown in several previous works \citep{wei2015submodularity, han2020replication, wang2021one}, many data utility functions using commonly used learning algorithms 
are close to submodular functions. 
While DeepSet-based utility learning models have already shown promising results in our experiment, DeepSets does not provide a mechanism to incorporate such prior knowledge. As another interesting line of future work, we would like to exploit the approximate submodularity of these kinds of data utility functions and use more fine-grained architectures or training algorithms for utility learning, e.g., submodular regularizations \cite{alievalearning}. 

\section{Conclusion}
This paper presents \AlgName as a unified framework for realizing task-driven, multi-purpose, model-agnostic data quality management. We theoretically analyzed the worst-case performance of existing data valuation-based DQM schemes and show that these approaches suffer unsatisfying performance guarantees. This sheds light on the empirical observations that existing data valuation-based DQM schemes exhibit significant performance variation over different datasets and tasks. Based on an extensive evaluation of the \AlgName over six types of DQM tasks and eight different datasets, we showed that \AlgName is more comprehensive and robust than the state-of-the-art DQM approaches with similar design goals.
For future work, we would like to further improve the scalability of the \AlgName as well as design utility learning models that are better aligned with the properties of data utility functions. 


\newpage

\bibliography{ref.bib}

\newpage

\appendix

\section{Proof of Theorem \ref{thm:nooptimal}}
\begin{customthm}{1}
For $n \ge 3$, there exists no linear heuristic $\M$ s.t.  $d(n, k) = {n \choose k}$ for every $k \in \{1, \ldots, n\}$. 
\end{customthm}
\begin{proof}
Suppose, for contradiction, that there exists a linear heuristic $\M$ s.t. $d(n, k) = {n \choose k}$ for all $k$. 
For a dataset $\dataset$ and utility function $U$, WLOG assume that the ranks (in non-ascending order) output by $\M$ in the Step 2 of Definition \ref{def:linearheuristic} is $(1, \ldots, n)$. 
Then it means 
\begin{align}
    &U(\{1\}) \ge U(S) \text{ for all } S \text{ s.t. } |S|=1,  \nonumber \\
    &U(\{1, 2\}) \ge U(S) \text{ for all } S \text{ s.t. } |S|=2, \nonumber \\
    &\ldots \nonumber \\
    &U(\{1, \ldots, n-1\}) \ge U(S) \text{ for all } S \text{ s.t. } |S|=n-1. \nonumber
\end{align}
We construct a simple counter example of $U$ to demonstrate such a $\M$ does not exist: let $n = 3$, we define $U$ as follows:
\begin{align}
    &U(\emptyset) = 0, \nonumber \\
    &U(\{1\}) = 7, U(\{2\}) = U(\{3\}) = 5, \nonumber \\
    &U(\{1, 2\}) = 9, U(\{1, 3\}) = 9, U(\{2, 3\}) = 10, \nonumber \\
    &U(\{1, 2, 3\}) = 10. \nonumber
\end{align}
To make $d(3, 1) = 3$, $\M$ must choose $1$ for $k=1$. However, for size-$2$ subsets, $\M$ can only choose between $\{1, 2\}$ and $\{1, 3\}$, whose utilities are both $9 < U(\{2, 3\})$. 
Therefore, $d(3, 2) = 2 < {3 \choose 2} = 3$. 
\end{proof}

\section{Proof of Theorem \ref{thm:symmetrybound}}
To formally state and prove Theorem \ref{thm:symmetrybound}, we introduce the formal definition of data type here. 
\begin{definition}
Given a dataset $\dataset$ and utility function $U$, if for all subset $S \subseteq \dataset \setminus \{i, j\}$, we have
$$
U(S \cup\{i\})=U(S \cup\{j\}),
$$
we say two data points $i$ and $j$ are of the same \emph{type}. 
\end{definition}
In other words, two data points are of the same type if they will be scored equally by every linear heuristic that satisfies Symmetry Axiom. 
Theorem \ref{thm:symmetrybound} essentially says that for all linear heuristic that will assign different scores to different types of data points, their domination numbers can be further upper bounded. 
We stress that this is a very mild assumption, especially when the space of the scores are continuous, which are the case for most of the existing data importance scoring mechanisms. 

To simplify the notations for set operations, we use $k \times \{D\}$ to denote a dataset that contains $k$ replicates of data point $D$, and we also denote the union of two data sets $S_1 \cup S_2 = S_1 + S_2$. 
The proof idea of Theorem \ref{thm:symmetrybound} is to construct a \emph{balanced} dataset that contains same amount of data points from the same types. If a linear heuristic $\M$ satisfies symmetry axiom, then $\M$ has to select data points of the same type when the target selection number is small, as all data points of the same type will receive the same scores. Of course, a dataset of only one type of data points will have nearly no utility. 

\begin{customthm}{2}[Restated]
If a linear heuristic $\M$ satisfies symmetry axiom and will always assign different scores for different types of data points, then the domination number $d(n, k)$ of $\M$ is upper bounded by $\floor{n/k}
\binom{\ceil{\frac{n}{\floor{n/k}}}}{k}$ for each $k \in \{1, \ldots, n\}$.
\end{customthm}


\begin{proof}
Suppose there are $c$ \emph{types} of data points: $D_1, \ldots, D_c$. Let $r = n \mod c$. We construct the dataset $\dataset$ that contains $\floor{n/c}$ data points for each of $D_1, \ldots, D_{n-r}$, and contains $\ceil{n/c}$ data points for each of $D_{n-r+1}, \ldots, D_{n}$. 
We construct utility function $U$ as follows: 
\begin{align}
    &U(\emptyset) = 0; \nonumber \\
    &U(i_1 \times \{ D_1 \} \ldots + i_c \times \{D_c\}) = 1, \nonumber
\end{align}
for every tuple of non-negative integers $(i_1, \ldots, i_c)$ s.t. $1 \le \sum_{j=1}^c i_j \le n$, 
except that 
\begin{align}
U( k \times \{ D_1 \} ) = \ldots = U( k \times \{ D_c \} ) = 0 \nonumber
\end{align}
for all $k \le \floor{\frac{n}{c}}$. 
This construction reflects the rationale that a dataset that only contains one type of data points (e.g. all of the same label) provide little information for the ML task. 

Since $\M$ satisfies symmetry axiom, we know that all data points of the same type will receive the same scores. 
Besides, we know that data points of different types will receive different scores. Therefore, when the target selection size $k \le 
\floor{\frac{n}{c}}$, $\M$ will return $k \times \{ D_j \}$, which has the worst utilities for subset at size $k$ and there are $(c-r){\floor{n/c} \choose k} + r{\ceil{n/c} \choose k}$ such subsets that only contains single types of data points. For each $k$, by taking the largest possible $c$ such that $k \le \floor{\frac{n}{c}}$, we obtain the desired bound. 
\end{proof}
We note that the upper bound is non-trivial for every $k \le n/2$. 
We also note the assumption that $\M$ always assigns different scores for different data types can be further relaxed as long as \emph{there exists} such a balanced dataset described in the proof that $\M$ assigns different scores for different data types.

\section{Proof of Theorem \ref{thm:shapleybound}}
Given a dataset $\dataset = \{1, \ldots, n\}$ and a submodular utility function $U$, the Shapley value is computed as 
\begin{equation}
v_{\text{shap}}(i) = \frac{1}{n}
\sum_{S \subseteq \dataset \setminus\{i\}} \frac{1}{{n-1 \choose |S|}}
\big[U(S\cup \{i\})-U(S)\big] 
\label{eq:shapley}
\end{equation}

\begin{customthm}{3}[Restated]
The domination number $d(n, k)$ of Shapley value is $1$ for every $n \ge 4$ and any $k \in \{1, \ldots, n\}$, even if we restrict the utility function $U$ to be submodular.
\end{customthm}
\begin{proof}
We first consider the case when $k \ge 3$. 

We construct an instance of a dataset $\dataset = \{1, \ldots, n\}$ and a submodular utility function $U$ as follows: 
\begin{align}
    &U(\emptyset) = 0; \nonumber \\
    &U(\{1\}) = U(\{2\}) = \ldots = U(\{k\}) = 7, 
U(\{i\}) = 5 \text{ for }i \ge k+1; \nonumber \\
    &U(S)=2|S|+5 \text{ for all }S \text{ s.t. } 2 \le |S| \le k-1; \nonumber \\
    &U(\{1, \ldots, k\}) = 2k+4,~~U(S)=2k+5 \text{ for all other }S\text{ s.t. } |S| = k; \nonumber \\
    &U(S)=2k+5  \text{ for all }S \text{ s.t. }|S| \ge k+1. \nonumber
\end{align}

We can compute Shapley value according to its definition in (\ref{eq:shapley}):
\begin{align}
v_{shap}(1) = \ldots = v_{shap}(k) 
&= \frac{1}{n} \left[7+\frac{2(k-1)+4(n-k)}{n-1} + 
2(k-3) + \frac{2{n-1 \choose k-1}-1}{ {n-1 \choose k-1} }\right] \nonumber \\
&= \frac{1}{n} \left[2k+3 + \frac{4n-2k-2}{n-1} -\frac{1}{{n-1 \choose k-1}}\right] \nonumber
\end{align}

\begin{align}
v_{shap}(k+1) = \ldots = v_{shap}(n) 
&= \frac{1}{n} \left[ 5 + \frac{2k+4(n-k-1)}{n-1} + 2(k-3) + \frac{1}{{n-1 \choose k-1}} \right] \nonumber \\
&= \frac{1}{n} \left[ 2k+1 + \frac{4n-2k-4}{n-1} - \frac{1}{{n-1 \choose k-1}} \right]  \nonumber
\end{align}

Since 
$$
v_{shap}(1) - v_{shap}(k+1) = 
\frac{1}{n} \left[
2 + \frac{2}{n-1} - \frac{1}{{n-1 \choose k-1}}
- \frac{1}{{n-1 \choose k}} \right] 
\ge \frac{2}{n(n-1)} >0,
$$
we know that $\M$ will always output $\{1, \ldots, k\}$, which achieves the lowest utility among all data subsets of size $k$. 
Therefore, Shapley value's domination number $d(n, k)=1$ for all $3 \le k \le n-1$.

We then consider the case when $k = 2$. The submodular data utility functions for the case of $k \ge 3$ can be easily adapted as follows: 
\begin{align}
    &U(\emptyset) = 0; \nonumber \\
    &U(\{1\}) = U(\{2\}) = 7, 
U(\{i\}) = 5 \text{ for }i \ge 3; \nonumber \\
    &U(\{1, 2\}) = 8,~~U(S)=9 \text{ for all other }S\text{ s.t. } |S| = 2; \nonumber \\
    &U(S)=9  \text{ for all }S \text{ s.t. }|S| \ge 3. \nonumber
\end{align}

The Shapley value is computed as follows:
\begin{align}
v_{shap}(1) = v_{shap}(2) 
&= \frac{1}{n} \left[7+\frac{1+4(n-2)}{n-1}\right] \nonumber \\
&= \frac{1}{n} \left[11-\frac{3}{n-1}\right] \nonumber
\end{align}

\begin{align}
v_{shap}(3) = \ldots = v_{shap}(n) 
&= \frac{1}{n} 
\left[ 5 + \frac{4+4(n-3)}{n-1}
+ \frac{2}{(n-1)(n-2)} \right]
\nonumber \\
&= \frac{1}{n} 
\left[ 9 - \frac{4}{n-1} + \frac{2}{(n-1)(n-2)} \right]
 \nonumber
\end{align}

Since 
$$
v_{shap}(1) - v_{shap}(3) = 
\frac{1}{n} \left[
2 + \frac{2}{n-1} - \frac{1}{{n-1 \choose k-1}}
- \frac{1}{{n-1 \choose k}} \right]
\ge \frac{2}{n(n-1)} >0,
$$
we know that $\M$ will always output $\{1, \ldots, 2\}$, which achieves the lowest utility among all data subsets of size $2$. 
Therefore, for Shapley value, $d(n, 2)=1$. 

Finally, we consider the case when $k = 1$. Similarly, we construct a submodular utility function as follows: 
\begin{align}
    &U(\emptyset) = 0; \nonumber \\
    &U(\{1\}) = 6, U(\{i\}) = 7 \text{ for }i \ge 2; \nonumber \\
    &U(\{1, i\}) = 11 \text{ for }i \ge 2, U(\{i, j\}) = 9 \text{ for }i, j \ge 2; \nonumber \\
    &U(S) = 11 \text{ for all }S \text{ s.t. }|S| \ge 3. \nonumber
\end{align}

The Shapley value is computed as follows:
$$
v_{shap}(1) = \frac{1}{n}[6+4+2] = \frac{12}{n}
$$
\begin{align}
v_{shap}(i) 
&= \frac{1}{n} \left[ 7 + \frac{5+2(n-2)}{n-1} + \frac{2(n-2)(n-3)}{(n-1)(n-2)} \right] \nonumber \\
&= \frac{1}{n}\left[11-\frac{1}{n-1}\right] \nonumber \\
&< \frac{12}{n} = v_{shap}(1). \nonumber
\end{align}
Therefore, Shapley value's domination number $d(n, k)=1$ for $k = 1$. 
\end{proof}

\section{Experiment Details and Results on More Datasets}

\subsection{Details of Figure 2}
In Figure 2 of the maintext, we showed the predicted vs true data utility values (test accuracy) for a synthetic dataset with logistic regression. 
For the synthetic data generation, we sample 200 data points from 50-dimensional standard Gaussian distribution. All of the 50-dimensional parameters are independently and uniformly drawn from $[-1, 1]$. Each data point is labeled by the sign of its vector's sum. 
The data utility model we use is a three-layer MLP (note that a set function can be represented by a function on $\{0, 1\}^n$ in a natural way). 

\subsection{Details of Figure 3}
In Figure 3 of the maintext, the tiny synthetic dataset is generated by sample data points from a 2-dimensional standard Gaussian distribution, where the mean vector of the Gaussian distribution is $(0.1, -0.1)$. Each data point is labeled by the sign of its vector's sum. We first sample 9 data points with positive label and 2 data points with negative label. We then replicate each of the two negatively labeled data points for two times. To simulate natural noise, we add Gaussian noise to the copied data vector with scale $10^{-5}$. By sampling and copying, we obtained 15 data points with natural redundancy. Since there are only 6 data points with negative label, they tend to be assigned with larger (and similar) importance scores by linear heuristics like Shapley value. Both Shapley and Least core thus rank negative points with higher importance. This means that when the target selection size is less than 6, the selected dataset will have only single kind of labels and no information about the other label class at all. As shown in Figure 3, both Shapley and Least core achieves trivial utility for the first 6 selected data points.

\subsection{Baseline Implementation}
For fair comparisons between \AlgName and baselines, we fix the total number of utility sampling as 4000 for \AlgName and baseline algorithms that require utility sampling, including Perm-SV, TMC-SV, G-SV, and LC. 
Following the settings in \cite{ghorbani2019data}, we set the performance tolerance in TMC-Shapley as $10^{-3}$. 
Following the settings in \cite{jia2019efficient}, we set $K=5$ for KNN-Shapley. 
We use CVXOPT\footnote{\url{https://cvxopt.org/}} library to solve the constrained minimization problem in the least core calculation. 
For influence function technique, we rank training data points according to their influences on the model loss over the validation data. The code is adapted from the PyTorch implementation of influence function on GitHub\footnote{\url{https://github.com/nimarb/pytorch_influence_functions}}. 
For TracIn technique, we only use the parameters in the last layer, following the settings in \cite{pruthi2020estimating}. 
We sample checkpoints for every 15 epochs. The implementation is adapted from the official GitHub repository\footnote{\url{https://github.com/frederick0329/TracIn}}.

\subsection{Details of Datasets Used in Section \ref{sec:eval}}

\paragraph{CIFAR-10 \citep{krizhevsky2009learning}.}
CIFAR-10 consists of 60,000 3-channel images in 10 classes (airplane, automobile, bird, cat, deer, dog, frog, horse, ship and truck). Each image is of size $32 \times 32$. 

\paragraph{MNIST \citep{lecun1998mnist}.}
MNIST consists of 70,000 handwritten digits. The images are $28 \times 28$ grayscale pixels. 

\paragraph{Dog vs. Cat \citep{dogcat}.} 
Dog vs. Cat dataset consists of 2000 images (1000 for `dog' and 1000 for 'cat') extracted from CIFAR-10 dataset. Each image is of size $32 \times 32$. 

\paragraph{Enron SPAM \citep{shams2013classifying}.}
Enron SPAM dataset consists of 2000 emails extracted from Enron corpus \cite{klimt2004enron}. The bag-of-words representation has 10714 dimensions. 

\paragraph{PubFig83 \citep{pinto2011scaling}.}
PubFig83 is a real-life dataset of 13,837 facial images for 83 individuals. Each image is resized to $32 \times 32$. 

\paragraph{Covid-CT \citep{zhao2020COVID-CT-Dataset}.}
The COVID-CT-Dataset has 746 CT images in total, containing 349 images from 216 COVID-19 patients and the rest of them are from healthy people. The dataset is separated into 543 training images and 203 test images. We resized each image to $32 \times 32$. 

\paragraph{UCI Adult Census \citep{uci-dataset}.}
The Adult dataset contains 48,842 records from the 1994 Census database. Each record has 14 attributes, including gender and race information. The task is to predict whether one's income exceeds \$50K/yr based on census data. 

\paragraph{COMPAS \citep{yoon2018machine}.}
We use a subset of the COMPAS dataset that contains 6172 data records used by the COMPAS algorithm in scoring defendants, along with their outcomes within two years of the decision, for criminal defendants in Broward County, Florida. 
Each data record has features including the number of priors, age, race, etc. 

\subsection{Implementation Details}
For the experiment of backdoor detection, data poisoning detection, noisy detection, and mislabel detection on the CIFAR-10 dataset, the CNN model we use has two convolutional layers. A max-pooling layer follows each with the ReLU as the activation function. 
For the experiment of Backdoor detection and noisy feature detection on the MNIST dataset, we use LeNet adapted from \cite{lecun1998gradient}, which has two convolutional layers, two max-pooling layers, and one fully-connected layer. 
For the experiment of data summarization, a large CNN model is adopted to train on the PubFig83 dataset, which has six convolutional layers, and each of them is followed by a batch normalization layer and a ReLU activation function. 
For the experiment on poisoning detection over the Dog vs. Cat dataset as well as the data summarization over the COVID-CT dataset, we use a small CNN model adapted from PyTorch tutorial\footnote{\url{https://pytorch.org/tutorials/beginner/blitz/cifar10_tutorial.html}}, which contains two convolutional layers, two max-pooling layers, and followed by three fully-connected layers. 
We use Adam optimizer with learning rate $10^{-3}$, mini-batch size 32 to train all of the models mentioned above for 30 epochs, except that we train LeNet for five epochs on MNIST. 
For the experiment of data biasing on the Adult dataset, we implement logistic regression in scikit-learn \cite{pedregosa2011scikit} and use the LibLinear solver. 
For the experiment of mislabeling detection on SPAM and data debiasing on COMPAS, we adopt SVM implementation from scikit-learn library \cite{pedregosa2011scikit} with RBF kernel. 

A DeepSets model is a set function $f(S) = \rho \left( \sum_{x \in S} \phi(x) \right)$ where both $\rho$ and $\phi$ are neural networks. In our experiment, both $\phi$ and $\rho$ networks have three fully-connected layers. For the COMPAS dataset, we set the number of neurons in every hidden layer and the dimension of set features (i.e., the output of $\phi$ network) to be 64. 
For all other datasets, we set the number of neurons and set dimension to be 128 . We use the Adam optimizer with learning rate $10^{-4}$, mini-batch size of 32, $\beta_1=0.9$, and $\beta_2=0.999$ to train all of the DeepSets utility models, for up to 20 epochs.

\subsection{Additional Results}
In this section, we present experiment details and results on more datasets corresponding to the applications introduced in the main body (see Section \ref{sec:eval}).

\subsubsection{Backdoor Attack}
We consider the two most popular types of backdoor attacks, namely the BadNets \cite{gu2017badnets} and the Trojan square trigger \cite{liu2017trojaning}. Those two attacks' major difference is the trigger itself, where BadNets adopts a white block trigger at the right corner, and Trojan attack adopts a square trigger. 

Here, we show the results of \AlgName and baseline techniques over detecting BadNets triggers on MNIST dataset. The poisoning rate is 0.25, and the target label is `0'. 
The performance of different DQM techniques is illustrated in Figure \ref{fig:appen_detection} I.(a) and I.(b). 
We can see that \AlgName outperforms all other methods in the detection rate and significantly reduces the attack accuracy after filtered out bad data points. 

\subsubsection{Data Poisoning Attack}
We discuss two popular types of clean-label data poisoning attacks. Feature collision attack \cite{shafahi2018poison} crafts poison images that collide with a target image in feature space, thus making it difficult for a model to discriminate between the two. 
Influence function-based poisoning attack \cite{koh2017understanding} identifies the most influential training data points for the target image and generates the adversarial training perturbation that causes the most increase in the loss on the target image. The Attack Success Rate is measured by the model's confidence on the prediction of poisoned data (with respect to the target label). 

Figure \ref{fig:appen_detection} II.(a) (b) show the results for influence function-based attack on Dog vs. Cat dataset, where 50 data points of class ‘cat’ are perturbed to increase the model loss on a ‘dog’ sample in the test set. As we can see, \AlgName is a more effective approach to detect poisoned data points than all other baselines.

\subsubsection{Noisy Feature}
\begin{figure}
    \centering%
	\includegraphics[width=0.99\linewidth]{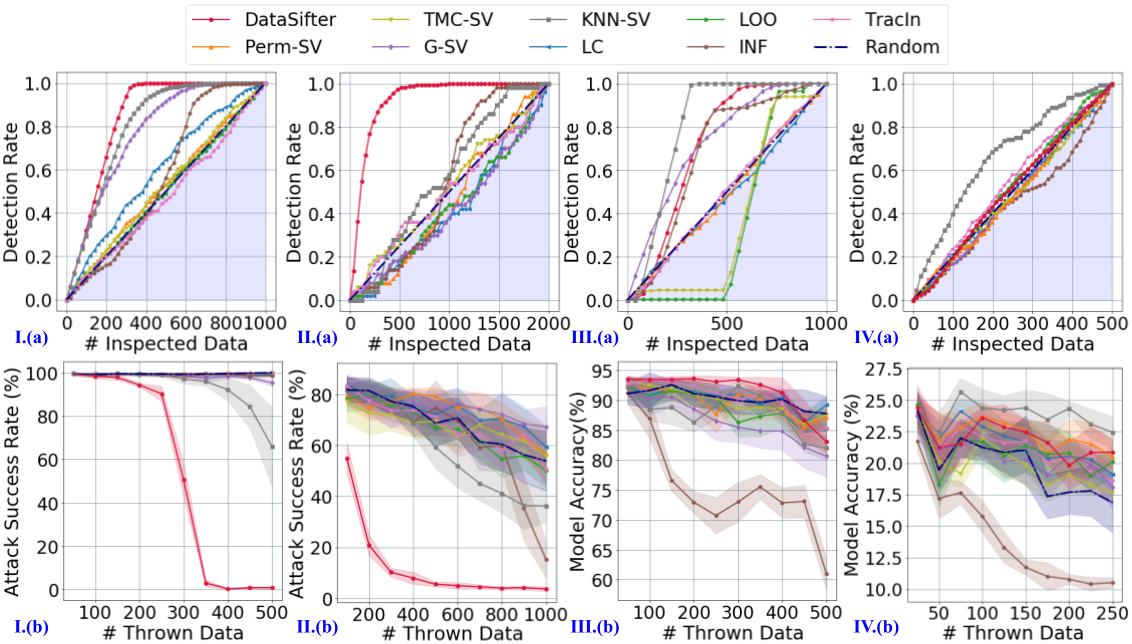}
	\caption{The experimental results and comparisons of the \AlgName under the case of filtering out harmful data (application I-IV). 
	The light blue region in each (a) graph represents the area that a method is no better than a random selection. For I.(b) and II.(b), we depict the Attack Success Rate (ASR), where a lower ASR indicates a more effective detection. 
	For III.(b) and IV.(b), we show the model test accuracy, where a higher accuracy means a better selection.}
    \label{fig:appen_detection}
\end{figure}

We follow the same evaluation method for noisy data detection as in Section \ref{sec:eval} with another setting: LeNet model trained on noise polluted MNIST. We randomly select 1000 data points and corrupt 25\% of them with white noise. 
As shown in Figure \ref{fig:appen_detection} III.(a) (b), we can see that although KNN-Shapley can achieve slightly better performance in detecting noisy data points, \AlgName still retains a higher performance for model accuracy. Besides, similar to the case for CIFAR10, we find that the KNN-SV approach only starts finding the noisy data points until filtering out a certain amount of clean data. 
This is mainly because all noisy data points are out-of-distribution (OOD), as shown in Figure \ref{fig:noisydataexample} (b). 
The mechanism of KNN-SV, however, tends to assign 0 values to OOD data points while assign negative values to clean data points that are in-distribution but have different labels from their neighbors. 
Figure \ref{fig:noisydataexample} (c) gives a visualization of the distribution of KNN-Shapley values.

\newcommand{\width}{0.3}

\begin{figure}
    \centering
    \begin{subfigure}[b]{0.29\textwidth}
        \includegraphics[width=\textwidth]{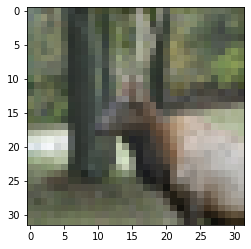}
        \caption{}
    \end{subfigure}
    \begin{subfigure}[b]{0.29\textwidth}
        \includegraphics[width=\textwidth]{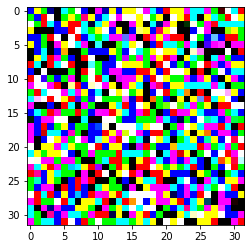}
        \caption{}
    \end{subfigure}
    \begin{subfigure}[b]{\width\textwidth}
        \includegraphics[width=\textwidth]{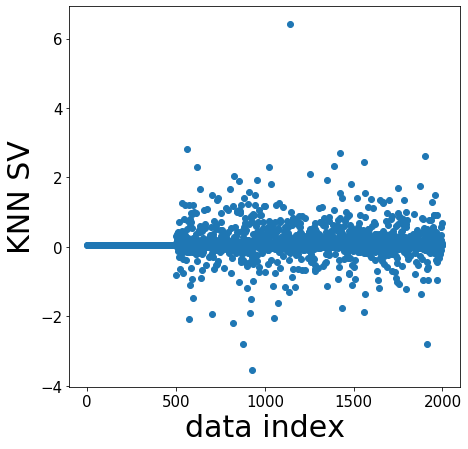}
        \caption{}
    \end{subfigure}
    \caption{(a) a normal image from CIFAR-10, (b) an example of noisy data image, (c) a sample of KNN-Shapley values, where data points with index $<500$ are noisy. A data point with a higher KNN-Shapley value is considered more important.}
    \label{fig:noisydataexample}
\end{figure}

\subsubsection{Mislabeled Data}
We conduct another experiment on noisy label detection: a small CNN model trained on 500 data points from the CIFAR-10 dataset. The noise flipping ratio is 25\%. The performance of mislabel detection is shown in Figure \ref{fig:appen_detection} IV.(a). As we can see, no DQM techniques are particularly effective in detecting mislabeled data for this task. Only KNN-SV achieves a slightly better performance than other approaches. We conjecture that the difficulty of mislabel detection on CIFAR-10 dataset is due to the following reason: since an oracle for detecting mislabeled data points can also be used to implement a classifier, the difficulty of mislabeling detection is at least as difficult as classification. A classifier directly trained on the 500 clean data points in this experiment, however, can only attain around 28\% test classification accuracy. 
Nevertheless, Figure \ref{fig:appen_detection} IV.(b) shows that \AlgName only achieves slightly worse model accuracy than KNN-SV after filtering out selected bad data points.

\begin{wrapfigure}{R}{0.65\textwidth}
    \centering
    \includegraphics[width=0.65\textwidth]{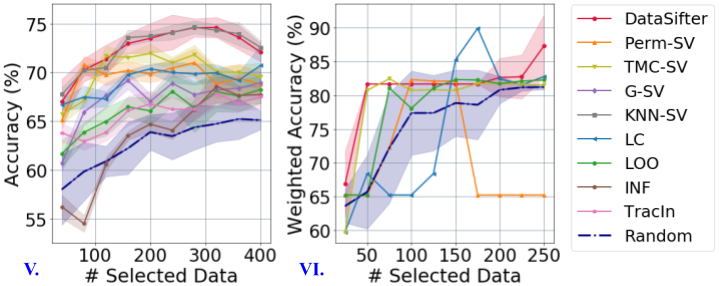}
    \caption{The experimental results and comparison of the \AlgName under the case of selecting high-quality data (application V and VI). 
    We depict the validation accuracy for both cases. A higher accuracy indicates a better performance. 
    }
    \label{fig:appen-selection}
\end{wrapfigure}

\subsubsection{Data Summarization}
As another setting for the data summarization application we consider, we use the patient CT images from COVID-CT dataset for a binary classification task, which aims to determine whether an individual is diagnosed with COVID-19 or not. The CNN model trained on the dataset achieves around 72\% classification accuracy. 
Figure \ref{fig:appen-selection} V. shows the results for selecting up to 400 data points with different DQM techniques. As we can see, \AlgName achieves the best model accuracies on the selected data points along with KNN-SV.

\subsubsection{Data Debiasing}
We introduce another data debiasing experiment on the criminal recidivism prediction (COMPAS) task, where races are considered as the sensitive attribute. 
The utility metric we adopted here is the average accuracy across different race groups. The learning algorithm we use is SVM with RBF kernel. 
Baselines including G-SV, KNN-SV, and Influence-based techniques are not applicable for this application due to the utility metric and learning algorithm we use. Figure \ref{fig:appen-selection} VI. shows the results for \AlgName and the remaining five baselines. We can see that \AlgName again achieves the top-tire performance. 


\subsubsection{Large Datasets}
\begin{figure}[ht!]
    \centering
    \includegraphics[width=0.65\textwidth]{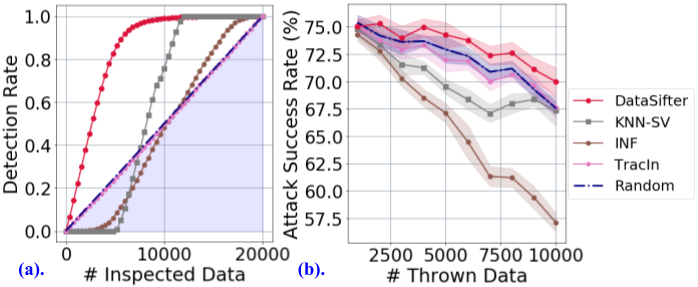}
    \caption{
    The experimental results and comparison of the \AlgName and baseline algorithms for detecting noisy data on larger datasets. 
    }
    \label{fig:appen-large}
\end{figure}

We follow the same protocol as in Section \ref{sec:largedata} for comparing the scalability between \AlgName and other baselines on a different setting: noisy data detection on a 20,000-size CIFAR-10 subset. The corruption ratio is 25\%. Again, for \AlgName, we use the learned utility model from Section \ref{sec:largedata} to select data points on the 20,000-size set. 
We remove the LOO, the Least core, and all the Shapley value-based approaches except KNN-SV from comparison, as they did not terminate in 24 hours for 4000 utility sampling on the 20,000-size set. As we can see from Figure \ref{fig:appen-large} (a) and (b), \AlgName significantly outperforms all other baseline techniques. The results demonstrate that although the utility sampling step could be expensive, the scalability of \AlgName can be boosted by the predictive power of the learned utility model. 

\end{document}


\maketitle

\appendix

\section{Proof of Theorem 1}
\begin{proof}
Suppose, for contradiction, that there exists a linear heuristic $\M$ s.t. $d(n, k) = {n \choose k}$ for all $k$. 
For a dataset $\dataset$ and utility function $U$, WLOG assume that the ranks output by $\M$ in the Step 2 of Definition \ref{def:linearheuristic} is $(1, \ldots, n)$. Then it means 
\begin{align}
    &U(\{1\}) \ge U(S) \text{ for all } S \text{ s.t. } |S|=1  \nonumber \\
    &U(\{1, 2\}) \ge U(S) \text{ for all } S \text{ s.t. } |S|=2 \nonumber \\
    &\ldots \nonumber \\
    &U(\{1, \ldots, n-1\}) \ge U(S) \text{ for all } S \text{ s.t. } |S|=n-1 \nonumber
\end{align}
A simple counter example: 
$$
U(\{ \}) = 0
$$

$$
U(\{1\}) = 7, 
U(\{2\}) = U(\{3\}) = 5
$$

$$
U(\{1, 2\}) = 9, U(\{1, 3\}) = 9, U(\{2, 3\}) = 10
$$
$\M$ must choose $1$ for $k=1$, then $\M(\I, 2) \le 9 < U(\{2, 3\})$. 
\end{proof}

\section{Proof of Theorem 2}
\begin{proof}
Suppose there are $c$ types of data, $D_1, \ldots, D_c$, and each has $\frac{n}{c}$ data points in $N$. 
We construct utility function $U$ as follows: 
$$
U(\{ \}) = 0
$$

$$
U(i_1 \times \{ D_1 \} \ldots + i_c \times \{D_c\}) = \sum_{j=1}^c j \times i_j
$$
for every tuple of non-negative integers $(i_1, \ldots, i_c)$ s.t. $1 \le \sum_{j=1}^c i_j \le n$, 
except that $U( r \times \{ D_1 \} ) = \ldots = U( r \times \{ D_c \} ) = 0$ for all $r \le \frac{n}{c}$. 
Since $\M$ satisfies symmetry axiom and $\M$ is non-trivial, we know that all $D_j$'s will receive the same score. Therefore, when $k \le \frac{n}{c}$, 
$\M(\I, k)$ will return $k \times \{ D_j \}$, which are the worst utilities for subset at size $k$ and has $c{n/c \choose k}$ such subsets.
\end{proof}

\section{Proof of Theorem 3}
\todo{for me: rewrite theorem statement}
\begin{proof}
We first consider the case when $k \ge 2$. 

We construct an instance of $n$ data points and submodular utility function as follows: 
$$
U(\{ \}) = 0
$$

$$
U(\{1\}) = U(\{2\}) = \ldots = U(\{k\}) = 7, 
U(\{i\}) = 5 \text{ for }i \ge k+1
$$

$$
U(S)=5 + 2|S| \text{ for all }S \text{ s.t. } 2 \le |S| \le k-1
$$

$$
U(\{1, \ldots, k\}) = 2k+4,~~U(S)=2k+5 \text{ for all other }S
$$

$$
U(S)=2k+5  \text{ for all }S \text{ s.t. }|S| \ge k+1
$$

We can compute Shapley value as follows:
\begin{align}
v_{shap}(1) = \ldots = v_{shap}(k) 
&= \frac{1}{n}[7+\frac{(k-1)2+(n-k)4}{n-1} + 
2(k-3) + \frac{2{n-1 \choose k-1}-1}{ {n-1 \choose k-1} }] \nonumber \\
&= \frac{1}{n}[2k+3 + \frac{4n-2k-2}{n-1} -\frac{1}{{n-1 \choose k-1}}] 
\end{align}

\begin{align}
v_{shap}(k+1) = \ldots = v_{shap}(n) 
&= \frac{1}{n} [ 5 + \frac{1}{n-1} (2k+4(n-k-1)) + 2(k-3) + \frac{1}{{n-1 \choose k-1}} ] \nonumber \\
&= \frac{1}{n} [ 2k+1 + \frac{4n-2k-4}{n-1} - \frac{1}{{n-1 \choose k-1}} ]
\end{align}

Since 
$$
v_{shap}(1) - v_{shap}(k+1) = 
2 + \frac{2}{n-1} - \frac{1}{{n-1 \choose k-1}}
- \frac{1}{{n-1 \choose k}} 
\ge \frac{2}{n-1} >0
$$
Therefore, Shapley value's domination number $d(n, k)=1$ for all $2 \le k \le n-1$. 

We now consider the case when $k = 1$. Similarly, we construct an instance of $n$ data points and submodular utility function as follows: 
$$
U(\{ \}) = 0
$$

$$
U(\{1\}) = 6, 
U(\{i\}) = 7 \text{ for }i \ge 2
$$

$$
U(\{1, i\}) = 11 \text{ for }i \ge 2, 
U(\{i, j\}) = 9 \text{ for }i, j \ge 2
$$

$$
U(S) = 11 \text{ for all }S \text{ s.t. }|S| \ge 3
$$

We can compute Shapley value as follows:
$$
v_{shap}(1) = \frac{1}{n}[6+4+2] = \frac{12}{n}
$$
\begin{align}
v_{shap}(i) 
&= \frac{1}{n}[ 7 + \frac{1}{n-1}[5+(n-2)2] + \frac{2}{(n-1)(n-2)}[(n-2)(n-3)] ] \\
&= \frac{1}{n}[11-\frac{1}{n-1}] \\
&< \frac{12}{n} = v_{shap}(1)
\end{align}
Therefore, Shapley value's domination number $d(n, k)=1$ for $k = 1$. 
\end{proof}

\section{Experiment Details and Results on More Datasets}

\subsection{Baseline Implementation}
For fair comparisons between \AlgName and baselines, we fix the total number of utility sampling as 4000 for \AlgName and baseline algorithms that require utility sampling, including Perm-SV, TMC-SV, G-SV, and LC. 
Following the settings in \cite{ghorbani2019data}, we set the performance tolerance in TMC-Shapley as $10^{-3}$. 
Following the settings in \cite{jia2019efficient}, we set $K=5$ for KNN-Shapley. 
We use CVXOPT\footnote{\url{https://cvxopt.org/}} library to solve the constrained minimization problem in Least core approximation. 
For influence function technique, we rank each training data point according to the estimated leave-one-out error over the validation data. The implementation is adapted from the PyTorch implementation of influence function on GitHub\footnote{\url{https://github.com/nimarb/pytorch_influence_functions}}. 
For TracIn, we only use the last layer, and we sample checkpoints for every 15 epochs. The implementation is adapted from the official GitHub repository\footnote{\url{https://github.com/frederick0329/TracIn}}.

\subsection{Details of Datasets Used in Section \ref{sec:eval}}

\paragraph{CIFAR-10.}
CIFAR-10 consists of 60,000 3-channel images in 10 classes (airplane, automobile, bird, cat, deer, dog, frog, horse, ship and truck). Each image is of size $32 \times 32$. 

\paragraph{MNIST.}
MNIST consists of 70,000 handwritten digits. The images are $28 \times 28$ grayscale pixels. 

\paragraph{Dog vs. Cat} 
Dog vs. Cat dataset consists of 2000 images (1000 for `dog' and 1000 for 'cat') extracted from CIFAR-10 dataset. Each image is of size $32 \times 32$. 

\paragraph{Enron SPAM}
Dog vs. Cat dataset consists of 2000 emails extracted from Enron corpus \cite{klimt2004enron}. The bag-of-words representation has 10714 dimensions. 

\paragraph{PubFig83.}
PubFig83 is a real-life dataset of 13,837 facial images for 83 individuals. Each image is resized to $32 \times 32$. 

\paragraph{Covid-CT.}
\todo{for Yi} Each image is resized to $32 \times 32$. 

\paragraph{UCI Adult Census.}
Adult dataset contains 48842 records from 1994 Census database. Each record has 14 attributes including gender and race information. The task is to predict whether income exceeds \$50K/yr based on census data.

\paragraph{COMPAS.}
We use a subset of COMPAS dataset contains 6172 data records used by the COMPAS algorithm in scoring defendants, along with their outcomes within 2 years of the decision, for criminal defendants in Broward County, Florida. 
Each data record has features including the number of priors, age, race, etc. 

\subsection{Implementation Details}
For the experiment of backdoor detection, data poisoning detection, noisy detection and mislabel detection on CIFAR-10 dataset, the CNN model we use has two convolutional layers, each is followed by a max pooling layer and a ReLU activation function. 
For the experiment of Backdoor detection and noisy feature detection on MNIST, we use LeNet adapted from \cite{lecun1998gradient}, which has two convolutional layers, two max pooling layers and one fully-connected layer. 
For the experiment of data summarization, a large CNN model is used to train on PubFig83, which has six convolutional layers, and each of them is followed by a batch normalization layer and a ReLU layer. 
For the experiment on poisoning detection on Dog vs Cat dataset as well as data summarization on COVID-CT dataset, we use a small CNN model adapted from PyTorch tutorial\footnote{\url{https://pytorch.org/tutorials/beginner/blitz/cifar10_tutorial.html}}, which contains has two convolutional layers, two max pooling layers and followed by three fully-connected layers. 
We use Adam optimizer with learning rate $10^{-3}$, mini-batch size 32 to train all of the aforementioned models for 30 epochs, except that we train LeNet for 5 epochs on MNIST. 
For the experiment of data biasing on Adult dataset, we implement logistic regression in scikit-learn \cite{pedregosa2011scikit} and use the LibLinear solver. 
For the experiment of mislabel detection on SPAM and data debiasing on COMPAS, we implement SVM with scikit-learn \cite{pedregosa2011scikit} and set the L2 regularization parameter to be $0.1$.

A DeepSets model is a function $f(S) = \rho ( \sum_{x \in S} \phi(x) )$ where both $\rho$ and $\phi$ are neural networks. In our experiment, both $\phi$ and $\rho$ networks have 3 fully-connected layers. For COMPAS dataset, we set the number of neurons in every hidden layer as well as the dimension of set features (i.e., the output of $\phi$ network) to be 64. For all other datasets, we set the number of neurons and set dimension to be 128 in each hidden layer. We use the Adam optimizer with learning rate $10^{-4}$, mini-batch size of 32, $\beta_1=0.9$, and $\beta_2=0.999$ to train all of the DeepSets utility models.

\subsection{Additional Results}
In this section, we present experiment details and results on more datasets corresponding to the applications introduced in the main body (see Section \ref{}).

\subsubsection{Backdoor Attack}
\todo{For Yi: add BRIEF descriptions for the two attacks.}

We consider another classic backdoor technique for this application: BadNets attack on MNIST dataset. The poisoning rate is 0.25, and the target label is `0'. 
The performance of different DQM techniques is illustrated in Figure \ref{fig:backdoor-poisoning-noisy} (a) and (d). 
We can see that \AlgName outperforms all other methods. 

\subsubsection{Data Poisoning Attack}
We discuss two popular types of clean-label data poisoning attacks. Feature collision attack \cite{shafahi2018poison} crafts poison images that collide with a target image in feature space, thus making it difficult for a model to discriminate between the two. 
Influence function-based poisoning attack \cite{koh2017understanding} identifies the most influential training data points for the target image, and generates the adversarial training perturbation that cause most increase in the loss on the target image. 

Figure \ref{fig:backdoor-poisoning-noisy} (b) (d) show the results for influence function-based attack on Dog vs Cat dataset, where 50 data points of class ‘cat’ are perturbed to increase the model loss on a ‘dog’ sample in the test set. As we can see, \AlgName is a more effective approach to detect poisoned data points than all other baselines.

\subsubsection{Noisy Feature}
We follow the same evaluation method for noisy data detection as in Section \ref{} on another setting: LeNet model trained on MNIST. The randomly select 1000 data points and corrupt 25\% of them with white noise. As shown in Figure \ref{fig:backdoor-poisoning-noisy} (c) (f), we can see that although KNN-Shapley can achieve slightly better performance in detecting noisy data points, \AlgName still retains a higher performance for model accuracy. Besides, similar to the case for CIFAR10 as shown in the main body, we find that the KNN-SV approach only starts finding the noisy data points until filtering out a certain amount of clean data. 
This is mainly because all noisy data points are out-of-distribution (OOD) as shown in Figure \ref{fig:noisydataexample} (b). 
The mechanism of KNN-SV tends to assign 0 values to OOD data points, while it will also assign negative values to clean data points that is in-distribution but have different labels from their neighbors, as illustrated in Figure \ref{fig:noisydataexample} (c).

\newcommand{\width}{0.3}
\begin{figure}
    \centering%
	\includegraphics[width=0.99\linewidth]{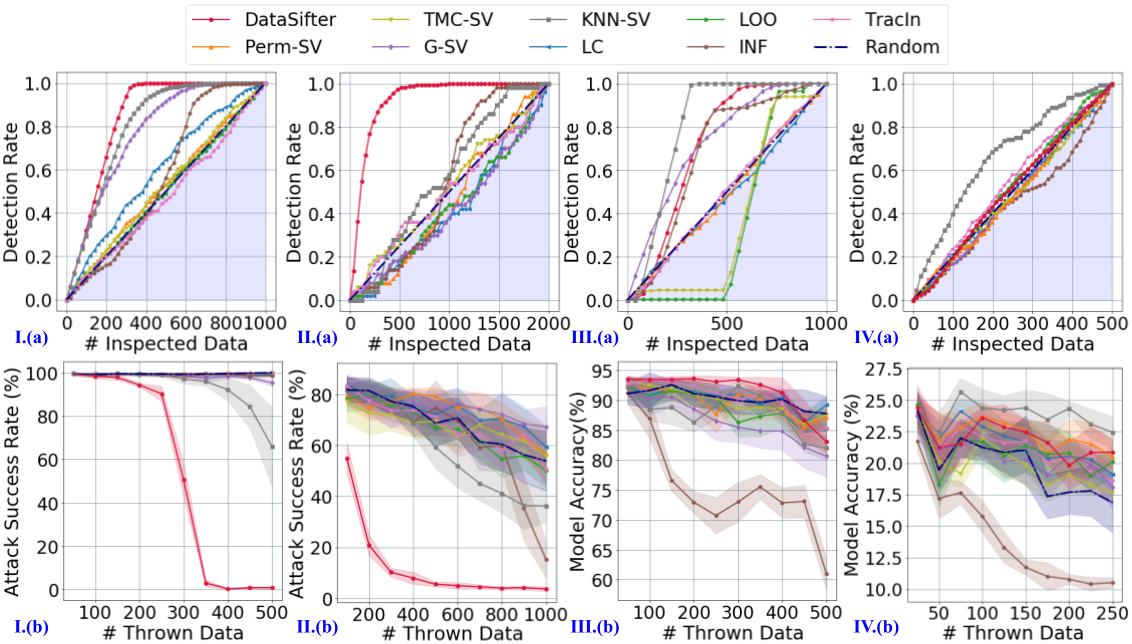}
	\caption{The experimental results and comparisons of the \AlgName under the case of filtering out harmful data (application I-IV). 
	The light blue region in each (a) graph represents the area that a method is no better than a random selection. For I.(b) and II.(b), we depict the Attack Success Rate (ASR), where a lower ASR indicates a more effective detection. For III.(b) and IV.(b), we show the model test accuracy, where a higher accuracy means a better selection.}
    \label{fig:appen_detection}
\end{figure}


\begin{figure}
    \centering
    \begin{subfigure}[b]{0.29\textwidth}
        \includegraphics[width=\textwidth]{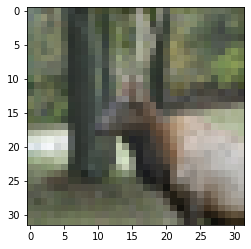}
        \caption{}
    \end{subfigure}
    \begin{subfigure}[b]{0.29\textwidth}
        \includegraphics[width=\textwidth]{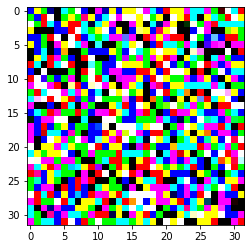}
        \caption{}
    \end{subfigure}
    \begin{subfigure}[b]{\width\textwidth}
        \includegraphics[width=\textwidth]{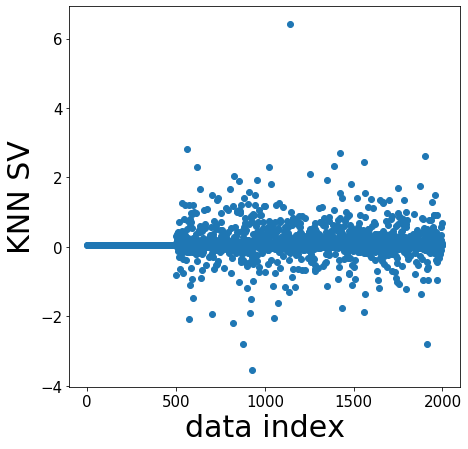}
        \caption{}
    \end{subfigure}
    \caption{(a) a normal image from CIFAR-10, (b) an example of noisy data image, (c) a sample of KNN-Shapley values, where data points with index $<500$ are noisy. A data point with higher KNN-Shapley value is considered more important. }
    \label{fig:noisydataexample}
\end{figure}

\subsubsection{Mislabeled Data}
We conduct another experiment on noisy label detection: a small CNN model trained on 500 data points from CIFAR-10 dataset. The noise flipping ratio is 25\%. The performance of mislabel detection is shown in Figure \ref{fig:appen_detection} IV (a). As we can see, no DQM techniques are extremely effective in detecting mislabeled data for this task, only KNN-SV achieves a slightly better performance than other approaches. We conjecture that the difficulty of this task is due to the following reason: since an oracle for detecting mislabeled data points can also be used to implement a classifier, the difficulty of mislabel detection is at least as difficult as classification. Since in this experiment, a classifier directly trained on the 500 clean data points have only around 28\% test accuracy, the best possible mislabel detection accuracy will also be relatively poor. 
Nevertheless, Figure \ref{fig:appen_detection} IV (b) shows that \AlgName only achieves slightly worse model accuracy than KNN-SV after filtering out selected bad data points. 
\todo{for Yi: help me to think about how to describe here}

\begin{wrapfigure}{R}{0.65\textwidth}
    \centering
    \includegraphics[width=0.65\textwidth]{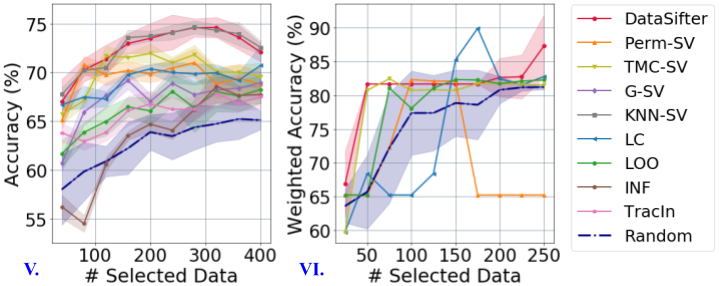}
    \caption{The experimental results and comparison of the \AlgName under the case of selecting high-quality data (application V and VI). 
    We depict the validation accuracy for both cases. A higher accuracy indicates a better performance. 
    }
    \label{fig:appen-selection}
\end{wrapfigure}

\subsubsection{Data Summarization}
As another setting for data summarization application we consider, we use the patient data in the COVID-CT dataset for the binary classification task of whether an individual is diagnoised with COVID19. Training CNN models result in around 72\% classification accuracy. 
Figure \ref{fig:appen-selection} V shows the results for selecting up to 400 data points with different DQM techniques. As we can see, \AlgName achieves the best model accuracies on the selected data points along with KNN-SV.

\subsubsection{Data Debiasing}
We introduce another data debiasing experiment on criminal recidivism prediction (COMPAS) task, where races are considered as the sensitive attribute. The fairness metric is the average accuracy across different race groups. Since the training algorithm we use is SVM, baselines including G-SV, KNN-SV, and Influence-based techniques are not applicable for this application. Figure \ref{fig:appen-selection} VI shows the results for \AlgName along with the remaining five baselines, where we can see that \AlgName again achieves the top-tire performance.

\begin{wrapfigure}{R}{0.65\textwidth}
    \centering
    \includegraphics[width=0.65\textwidth]{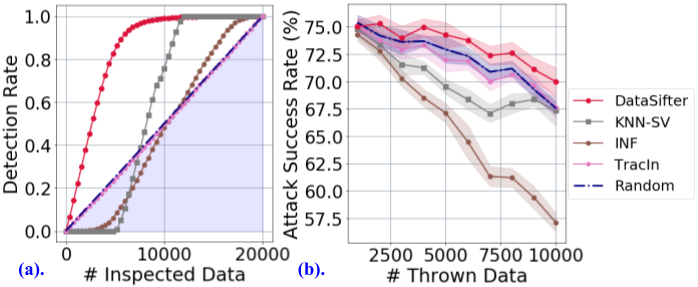}
    \caption{
    The experimental results and comparison of the \AlgName and baseline algorithms for detecting noisy data on larger datasets. 
    }
    \label{fig:appen-large}
\end{wrapfigure}

\subsubsection{Large Datasets}
We follow the same protocol as in Section \ref{} for comparing the scalability between \AlgName and other baselines on a different setting: noisy data detection on a 20,000-size CIFAR-10 subset. The corruption ratio is 25\%. Again, for \AlgName we use the learned utility model from Section \ref{} to select data points on 20,000 data points. 
We remove the LOO, the Least core, and all the Shapley value-based approaches except KNN-SV from comparison, as they did not terminate in 24 hours. As we can see from Figure \ref{fig:appen-large} (a) and (b), \AlgName significantly outperforms all other baseline techniques. The results demonstrate that although the utility sampling step could be expensive, the scalability of \AlgName can be boosted by the predictive power of the learned utility model. 

\bibliography{ref.bib}